# On the Limitation of Convolutional Neural Networks in Recognizing Negative Images

Hossein Hosseini, Baicen Xiao, Mayoore Jaiswal and Radha Poovendran
Network Security Lab (NSL), Department of Electrical Engineering, University of Washington, Seattle, WA
{hosseinh, bcxiao, mayoore, rp3}@uw.edu

*Abstract*—Convolutional Neural Networks (CNNs) have achieved state-of-the-art performance on a variety of computer vision tasks, particularly visual classification problems, where new algorithms reported to achieve or even surpass the human performance. In this paper, we examine whether CNNs are capable of learning the semantics of training data. To this end, we evaluate CNNs on negative images, since they share the same structure and semantics as regular images and humans can classify them correctly. Our experimental results indicate that when training on regular images and testing on negative images, the model accuracy is significantly lower than when it is tested on regular images. This leads us to the conjecture that current training methods do not effectively train models to generalize the concepts. We then introduce the notion of *semantic adversarial examples* – transformed inputs that semantically represent the same objects, but the model does not classify them correctly – and present negative images as one class of such inputs.

## I. INTRODUCTION

Deep Neural Networks (DNNs) have transformed the machine learning field and are now widely used in many applications. One of the fields that has benefited the most from deep learning is computer vision, where Convolutional Neural Networks (CNNs) have achieved state-of-the-art results on variety of problems [1], [2], [3]. For instance, new algorithms for image classification are reported to reach or even surpass the human performance [4], [5], [6].

Several recent papers have also hypothesized that CNNs develop an understanding about objects based on the training data, as such that they are even able to generate new images [7], [8]. Humans however have an incredible capability in recognizing unfamiliar objects, by identifying their important features, mainly their shapes [9]. They can also identify objects in various forms such as different scales, orientations, colors or brightness. Therefore, it remains to be seen how CNNs compare to humans in terms of "semantic generalization."

In this paper, we evaluate the performance of CNNs on *negative images*. A negative is referred to an image with reversed brightness, i.e., the lightest parts appear the darkest and the darkest parts appear lightest. Unlike typical transformations used for training or testing machine learning models, image complementing causes a large pixel-wise perturbation to the original images. It however maintains the structure (e.g., edges) and semantics of the images, as negative images are often easily recognizable by humans. To the best of our

This work was supported by ONR grants N00014-14-1-0029 and N00014-16-1-2710, ARO grant W911NF-16-1-0485 and NSF grant CNS-1446866.

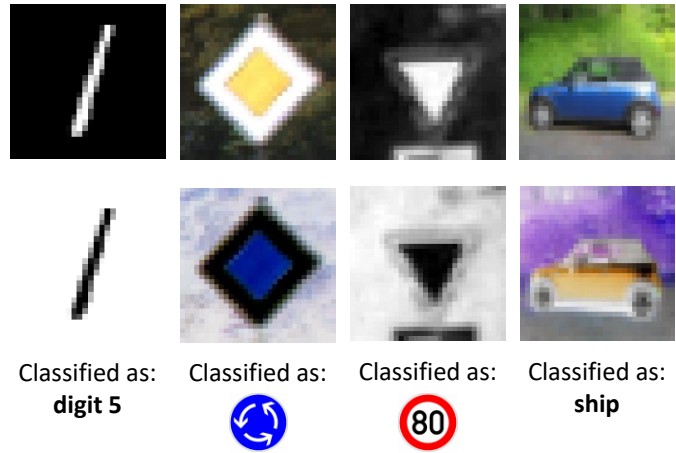

Fig. 1: Examples of original images (top) and negative images (bottom) from datasets MNIST, GTSRB (color version), GTSRB (gray-scale version) and CIFAR. CNNs trained with regular images significantly underperform when tested with negative images.

knowledge, our paper is the first work to study negative images as a form of data augmentation.

We conduct extensive experiments on widely used CNN architectures and standard image datasets. The results are presented for LeNet-5 network [10] trained on MNIST dataset and modified VGG networks [3] trained on color and gray-scale versions of German Traffic Sign Recognition Benchmark (GTSRB) and also CIFAR-10 dataset. We study the role of data augmentation, network depth, diversity of training data and complexity of features in the capability of model in recognizing negative images. We show that when training on regular images and testing on negative images, the accuracy of CNNs is significantly lower than when they are tested with regular images. Specifically, we found that the accuracy on negative images is relatively good, only if there is significant diversity within the training data.

Our results indicate that neural networks underperform when test data is not exactly distributed as the training data, a scenario that frequently happens in practice. We argue that this is due to the fact that current training methods push the network to memorize the inputs [11] and thus put the burden on the training data to be rich. As a result, the model does not effectively learn the structures of the objects and cannot semantically distinguish between classes. We also suggest that merely computing the accuracy on the test data, that is distributed similarly as the training data, is not representative of the behavior of machine learning models in the wild. Therefore, our results call for better training methods and also more meaningful performance metrics.

TABLE I: LeNet−5 and Modified VGG (MVGG) architectures used in experiments. In table, "conv $x \times x \times y$" denotes a convolutional layer with $y$ filters of kernel size $x \times x$, "max pool $x \times x$" denotes a max pooling layer with $x \times x$ filters, "FC $x$" is a fully-connected layer with $x$ rectified linear units, and "softmax" is the softmax layer.

| LeNet−5 | MVGG−5 | MVGG−6 | MVGG−7 | MVGG−8 | MVGG−9 |
|---|---|---|---|---|---|
| conv $5 \times 5 \times 6$ | conv $3 \times 3 \times 16$ | conv $3 \times 3 \times 16$ | conv $3 \times 3 \times 16$ | conv $3 \times 3 \times 16$ | conv $3 \times 3 \times 16$ |
| max pool $2 \times 2$ | conv $3 \times 3 \times 16$ | conv $3 \times 3 \times 16$ | conv $3 \times 3 \times 16$ | conv $3 \times 3 \times 16$ | conv $3 \times 3 \times 16$ |
| conv $5 \times 5 \times 16$ | max pool $2 \times 2$ | max pool $2 \times 2$ | max pool $2 \times 2$ | max pool $2 \times 2$ | max pool $2 \times 2$ |
| max pool $2 \times 2$ | conv $3 \times 3 \times 48$ | conv $3 \times 3 \times 32$ | conv $3 \times 3 \times 32$ | conv $3 \times 3 \times 32$ | conv $3 \times 3 \times 32$ |
| conv $5 \times 5 \times 120$ | max pool $2 \times 2$ | conv $3 \times 3 \times 48$ | conv $3 \times 3 \times 32$ | conv $3 \times 3 \times 32$ | conv $3 \times 3 \times 32$ |
| FC 84 | FC 128 | max pool $2 \times 2$ | max pool $2 \times 2$ | max pool $2 \times 2$ | max pool $2 \times 2$ |
| FC 10 | FC 10 | FC 128 | conv $3 \times 3 \times 48$ | conv $3 \times 3 \times 48$ | conv $3 \times 3 \times 48$ |
| softmax | softmax | FC 10 | max pool $2 \times 2$ | conv $3 \times 3 \times 48$ | conv $3 \times 3 \times 48$ |
| | | softmax | FC 128 | max pool $2 \times 2$ | max pool $2 \times 2$ |
| | | | FC 10 | FC 128 | conv $3 \times 3 \times 64$ |
| | | | softmax | FC 10 | max pool $4 \times 4$ |
| | | | | softmax | FC 128 |
| | | | | | FC 10 |
| | | | | | softmax |

Moreover, the fragility of machine learning models to transformed inputs has a security implication. An adversary, who has no access to the learning system, can generate transformed inputs which are semantically representing the same objects, yet the model does not correctly classify them. We call such transformed inputs as *semantic adversarial examples* and propose the image complementing as one such transformation.

The rest of this paper is organized as follows. Section II reviews related literature and Section III gives the definition of negative images. Experimental results are provided in Section IV. Section V discusses the limitation of CNNs in recognizing transformed inputs and its security implications. Section VI concludes the paper.

## II. RELATED WORKS

Several papers have proposed various forms of data augmentation as an approach for improving the generalization capability of machine learning models. Data augmentation is achieved by subjecting the training data to transformations such as rotation, flipping, spatial shifting, color perturbation, quantization, adding noise, resizing/scaling, or other transformations that realistically represents the distribution of dataset [2], [12], [13], [14]. Image transformations are also used to evaluate the performance of deployed CNNs [12].

However, many of these transformations only introduce slight variations to the input, and hence are limited in evaluating the generalization capabilities of trained models. In contrast, image complementing yields images that are semantically similar to original images, but very different in pixel space. Therefore, they can be used as a check for how much the model can semantically generalize.

Generalizing to images with different distributions is also a subject of transfer learning [15], [16], [17] and domain adaptation [18], [19], [20]. The goal is to use models and features learned on one dataset/domain for another dataset/domain with inimal fine-tuning [21]. To the best of our knowledge, generalizing to negative images is not studied in transfer learning or domain adaptation literature.

## III. NEGATIVE IMAGES

In this paper, we examine whether CNNs are capable of learning the semantics of training data. To this end, we evaluate CNNs on negative images. A negative image is defined as the image complement of the original image, in which the light pixels appear dark and vice versa. Let $X$ be an image and $X_{i,n} \in [0, 1]$ be the $i$-th pixel in the $n$-th color channel. The negative image is defined as $X^*$, where $X^*_{i,n} = 1 - X_{i,n}$. Image complementing is a simple transformation that preserves the structure (e.g., edges) and semantics of the image and typically does not impact the human perception of the object. Figure 1 shows examples of original and negative images. In the rest of the paper, we refer to the original images as regular images.

## IV. EXPERIMENTAL RESULTS

We conducted the experiments on datasets MNIST [22], CIFAR-10 [23] and German Traffic Sign Recognition Benchmark (GTSRB) [24]. We used CNN architectures LeNet-5 [10] and modified versions of VGG [3]. In the following, we describe the datasets and CNN architectures in details and provide the experimental results.

### A. Datasets

**MNIST.** MNIST is an image dataset of handwritten digits consisting of $50,000$ training samples, $10,000$ validation samples, and $10,000$ test samples [22]. Images are gray-scale and of size $28 \times 28$ pixels. MNIST has 10 classes representing digits from 0 to 9.

**CIFAR-10.** CIFAR-10 dataset consists of natural color images of size $32 \times 32$ pixels in 10 classes of airplane, automobile, bird, cat, deer, dog, frog, horse, ship, and truck [23]. The training and test sets contain $50,000$ and $10,000$ images, respectively, and we hold out $5,000$ training images as a validation dataset.

TABLE II: The test accuracy of different classifiers on regular and negative images of various image datasets. The CNN architectures are trained on regular training data.

| Classifier | Dataset | Accuracy on regular images | Accuracy on negative images |
|---|---|---|---|
| LeNet−5 | MNIST | 99.21% | 34.65% |
| LeNet−5 | MNIST with data augmentation | 99.25% | 12.38% |
| MVGG−5 | CIFAR-10 | 78.24% | 38.55% |
| MVGG−6 | CIFAR-10 | 80.86% | 41.66% |
| MVGG−7 | CIFAR-10 | 82.78% | 45.51% |
| MVGG−8 | CIFAR-10 | 83.17% | 46.16% |
| MVGG−9 | CIFAR-10 | 84.01% | 47.88% |
| MVGG−8 | GTSRB-color | 98.54% | 12.66% |
| MVGG−8 | GTSRB-gray | 98.12% | 12.29% |
| Human | GTSRB-color | 98.48% | 97.31% |
| Human | GTSRB-gray | 98.00% | 96.41% |

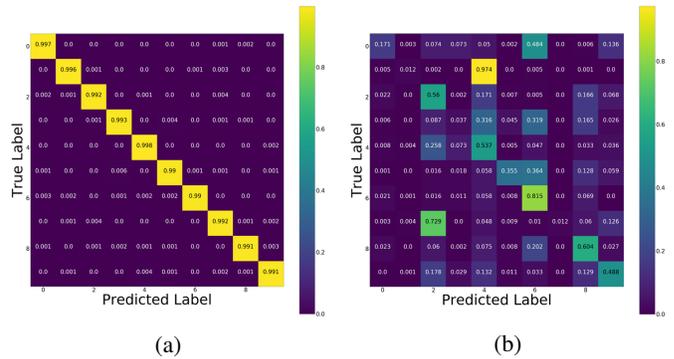

Fig. 2: Confusion matrices of LeNet−5 model trained on MNIST training images and tested on a) regular test images and b) negative test images. The value of entry $(i, j)$ indicates the probability that an image of class $i$ is classified into class $j$. The values in each row sum to $1$.

**GTSRB.** GTSRB is a real-world traffic sign dataset [24]. It consists of color images of $43$ traffic signs with $39,209$ training samples and $12,630$ test samples. For our experiments, we resized all images to $32 \times 32$ pixels and used $20\%$ of training images from each class as validation to tune the hyperparameters. We also generated the gray-scale version of the dataset by converting each color image to gray-scale. We call the color and gray-scale versions of GTSRB as GTSRB-color and GTSRB-gray, respectively. The results are provided on both versions of the dataset.

### B. CNN Architectures

**LeNet-5 Architecture.** The LeNet-5 [10] architecture is described in Table I. The architecture consists of three convolutional layers followed by two fully-connected layers. The classification is made by a softmax layer. This network was trained on MNIST dataset.

**Modified VGG Architectures.** We modified the VGG [3] architecture to work with input images of size $32 \times 32$. We used architectures with different depth, which are outlined in Table I. Similar to VGG−16, the convolution kernels have a fixed size of $3 \times 3$. The number of convolutional layers increases as the depth of the CNN increases, and all architectures have two fully connected layers before the softmax layer. The modified VGG architectures were trained on CIFAR-10 and GTSRB datasets.

### C. Accuracy of CNNs on Negative Images

To evaluate the ability of CNN architectures to capture the semantics of training data, we train them on regular training images and then test on both regular and negative test images. Negative images have the same structure as regular images. Specifically, for MNIST, GTSRB and CIFAR-10 datasets, the classes are very distinct and image complementing does not change the ground-truth label. This however may not be true for datasets with very high number of closely related classes, such as ImageNet dataset [25].

Table II shows the accuracies of different classifiers on regular and negative test images. As can be seen, CNNs yield high accuracy on regular images, which are drawn from the same distribution as training data. They however significantly underperform when testing on negative images.

The ability of a network to generalize to negative images seems to depend on the complexity of the features and the diversity of the training data. For MNIST dataset, different classes can be distinguished only by the edge information, regardless of the background and foreground pixel values. Figure 2 provides the confusion matrices for predicted labels of both regular and negative images, for a LeNet-5 architecture trained on MNIST. As expected, the confusion matrix of the model on regular test images exhibits a strong diagonal structure. However, based on the confusion matrix of the model on negative test images, the model seems to only partially recognize the digits in negative images, and it is biased towards predicting digits $4$ and $6$.

The images in CIFAR-10 dataset have more complex features than images in MNIST. However, the training images of CIFAR-10 contain a lot of diversity in object colors. For instance, the bird images represent birds with different colors, which causes the model to learn to be somewhat invariant to the object color. Therefore, the intra-class diversity of the CIFAR-10 dataset helps the model to generalize better to negative images compared to the MNIST dataset.

In contrast, GTSRB dataset has less complex features than CIFAR-10, but higher number of classes. Moreover, the samples within classes are highly correlated, which causes the model to hardly generalize to images with different color or gray values. For GTSRB dataset, a random classifier yields an accuracy of about $3.76\%$ on test images[1]. The models trained on GTSRB-color and GTSRB-gray datasets yield similar accuracy of about $12.5\%$ on negative test images, which is more than the random classifier, but significantly lower than the accuracy on regular test images. Moreover, our results indicate that merely using color datasets does not automatically help the semantic generalization. The model rather needs diverse images so that it learns the invariant features of each of the classes.

---
[1]There is an imbalance in number of samples from different classes of GTSRB dataset. We found that a random classifier yields on average $3.76\%$ accuracy on test images.

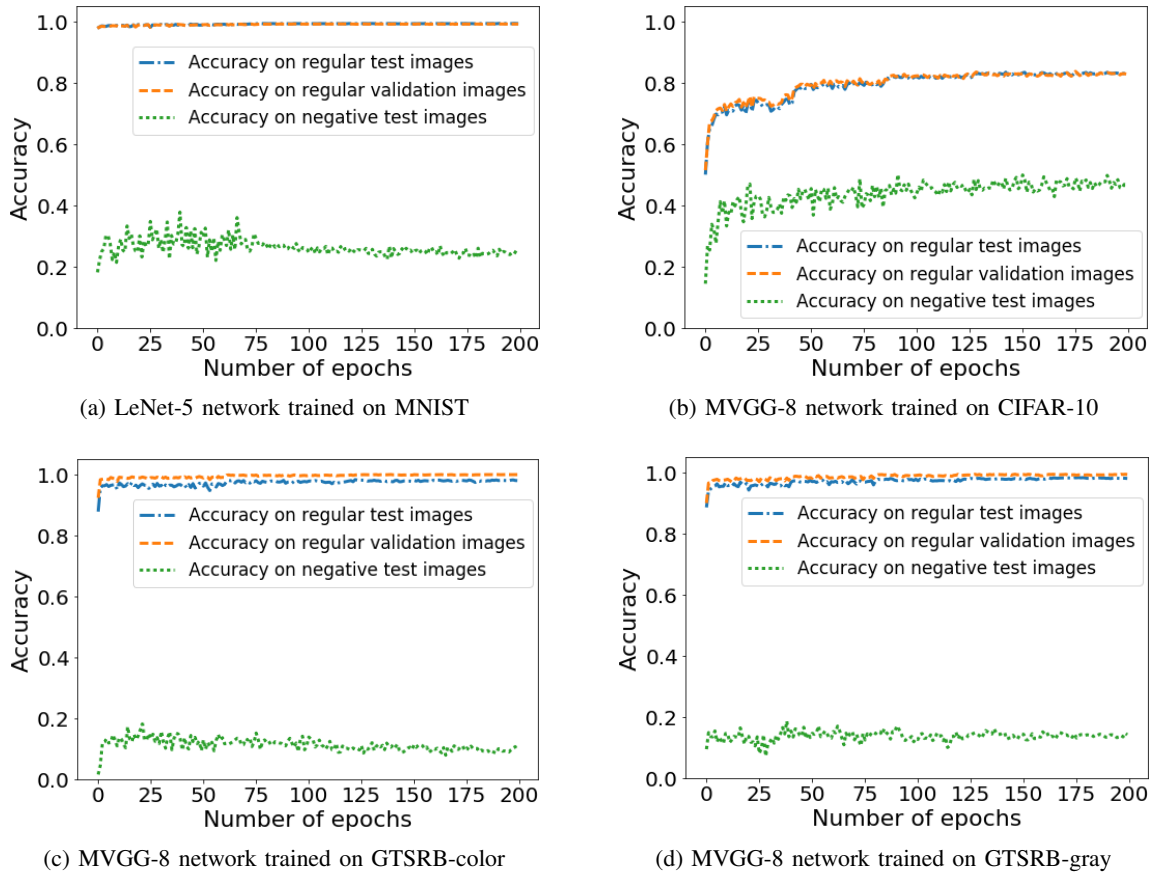

Fig. 3: Accuracy of CNN models versus number of training epochs. Throughout training, accuracy on regular test and validation images closely match. However, the accuracy on negative test images remains significantly lower. (Best viewed in color)

Figure 3 plots the accuracy on regular validation images, regular test images, and negative test images versus the epoch number for different CNNs. As expected, the accuracy on regular test and validation images closely match throughout the training. However, as training proceeds, the model seems to overfit to distribution of the training data and thus performs worse on negative images.

**Effect of data augmentation.** A common method to help machine learning models to generalize better is data augmentation [2], which is the process of transforming the training data in a manner that the labels are preserved. To evaluate the effect of data augmentation on the accuracy on negative images, we augmented the MNIST training data by computing image translations and reflections. As shown in Table II, the accuracy on regular test images did not change much, but surprisingly the accuracy on negative test images significantly dropped. We repeated the experiments several times and consistently obtained similar results.

The results demonstrate that data augmentation causes the model to "semantically" overfit to the distribution of the training data, leading to poor accuracy on negative images. In other words, simple data transformations do not semantically augment the training data and the potential increase in test accuracy seems spurious. Therefore, we argue that simply evaluating the machine learning models based on the test accuracy can be misleading.

**Effect of architecture depth.** To evaluate the effect of depth in network performance, we used modified VGG networks with different number of layers, trained on CIFAR-10 dataset. As shown in Table II, CNNs with more number of layers yield higher accuracy on both regular and negative test images. This implies that deeper networks can better capture the important features of the training images and thus have more generalization capability.

**Training CNN models with negative images.** We examine that, when training on negative images, how fast the model can learn to recognize them. For this, we perform two experiments: 1) fine-tuning the CNN model trained on regular images with different number of negative images, and 2) training the CNN model from scratch with different number of negative images. Figure 4 plots the results for MNIST and GTSRB datasets.

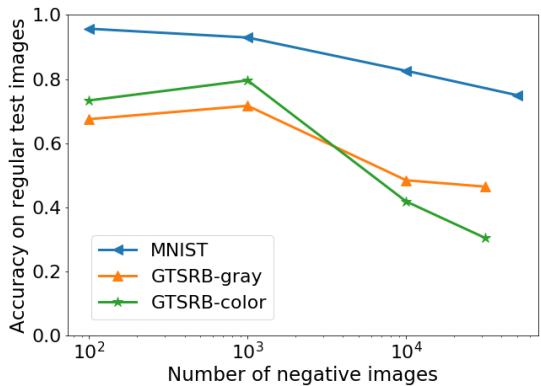
(a) Accuracy of CNN models, trained on regular images and fine-tuned with negative images.

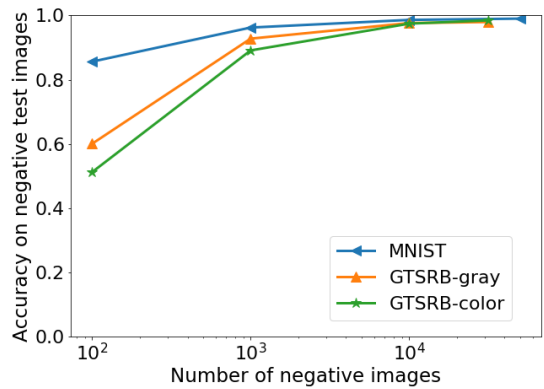
(b) Accuracy of CNN models, trained on regular images and fine-tuned with negative images.

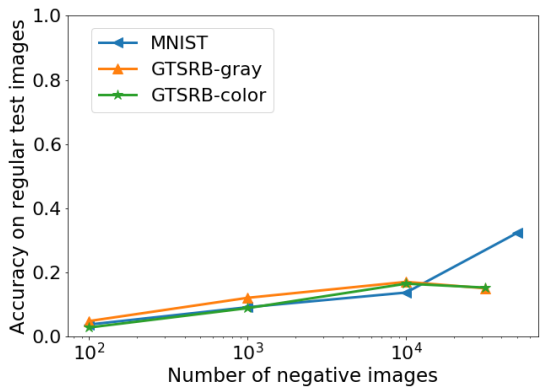
(c) Accuracy of CNN models trained with negative images from scratch.

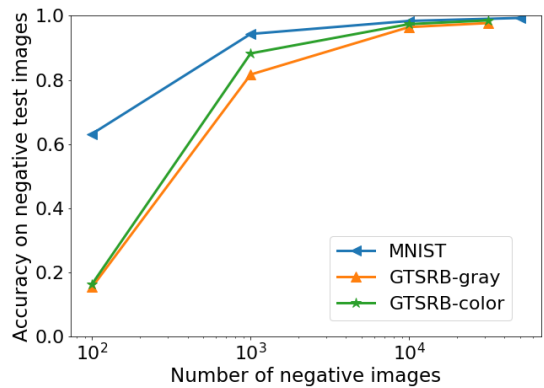
(d) Accuracy of CNN models trained with negative images from scratch.

Fig. 4: The accuracy on regular and negative test images for CNN models trained on different number of negative training images. In (a-b), the model is trained on regular training images and fine-tuned with negative images, whereas in (c-d), the model is trained with negative images from scratch.

For fine-tuned models, since we only retrain the model on negative images, the model's accuracy decreases on regular test images. Also, as expected, by training on more negative images, the accuracy on negative test images increases. For models trained from scratch on negative images, with more images, the model's accuracy increases on both regular and negative images; though, the accuracy remains low for regular images. Moreover, the figure shows when testing on negative images, fine-tuning a model which has been trained on regular images yields better results. However, as we train the models with more negative images, the advantage diminishes.

### D. Human Perception of Negative Images

In this subsection, we present the results of human accuracy in recognizing negative images of GTSRB-gray and GTSRB-color datasets. The experiment is described as follows. We provided 10 people with samples from GTSRB test images, containing regular color images, negative color images, regular gray-scale images and negative gray-scale images. We also provided the participants with representative images from GTSRB classes (shown in Figure 5a). We then asked them to map each image to one of the dataset classes. The participants were unaware that the samples include negative images of test data.

The results are provided in Table II. As can be seen, the performance is the best on color images and worst on negative gray-scale images. However, the reduction in accuracy from regular images to negative images is very small. Specifically, the accuracy decreases only about 1% when testing on negatives of color images and only about 1.5% when testing on negatives of gray-scale images. Note that, when asking humans to annotate the images, the question is not "What object the image represents?", but rather "Which class the image belongs to?". In other words, similar to the experiments on machine learning models, we conducted a *closed-world experiment*, where we required the participants to link each image to one of the classes.

Figure 5 illustrates the human reasoning for mapping images to classes of GTSRB dataset. The figure represents images of shapes diamond, inverted triangle and octagon. These images are fabricated, i.e., they are not chosen from regular or negative images of the GTSRB dataset. Note that, although the fabricated images do not exactly look like any of the GTSRB representative images, humans can easily associate each one of them with one traffic sign, essentially because there are only three signs with diamond, inverted triangle and octagon shapes. In other words, given the image, humans realize that the color information is lost; they thus look for the most important feature in the image, which can represent a class the best and distinguish the image the most from other classes. This can be also attributed to the fact that humans have a strong "shape bias," i.e., they prefer to categorize objects according to shape rather than color [9].

In contrast, neural network models learn representations based on the training data. When testing with a new sample, the model passes it through the pre-defined filters and maps the image to a class that it resembles the most. The problem however is that, in inference time, the inputs can be very different from the training data. Therefore, the pre-defined representations are not sufficient to semantically distinguish between the images from different classes. Instead, just as what humans do, the model needs to look for specific features that are most representative for the given image. This can be potentially done using architectures such as Matching Networks, which are designed for the task of associating a new sample with a small set of training data [26].

## V. DISCUSSION

In this section, we discuss the fragility of CNNs to transformed inputs and examine its security implications.

### A. Fragility of CNNs to Transformed Inputs

It has been shown that the effective capacity of neural networks is sufficient for memorizing the entire training dataset [11]. As a result, neural network classifiers generally correctly classify the training samples with very high confidence. Besides, the network loss function vary smoothly around the input samples, i.e., a *randomly* perturbed sample is likely to be classified into the same class as the regular sample [27]. Since test samples are typically collected from the same distribution as training samples, the test data points occur mostly in vicinity of the training points. Therefore, with the availability of large datasets, it is likely that the network can associate each test sample with one or several training samples from the same class and thus achieve high test accuracy.

However, if a transformed sample has a large pixel-wise difference to the original sample, the network might not be able to correctly classify it. Of course, for a particular transformation, we can train the model also on the transformed data to get high accuracy on them. However, relying on training data to cover all aspects of possible novelties at the inference time poses a fundamental limit in adaptation of machine learning systems in real-world applications. While many computer vision problems are data rich, for some critical applications, e.g., training driveless cars, gathering diverse training data is costly. To address this issue, several papers have proposed methods, such as one-shot or zero-shot learning, to learn with small datasets [28], [29], [30], [26], In essence, learning to reason is the key to "semantical generalization" and can compensate for the lack of diversity in training data.

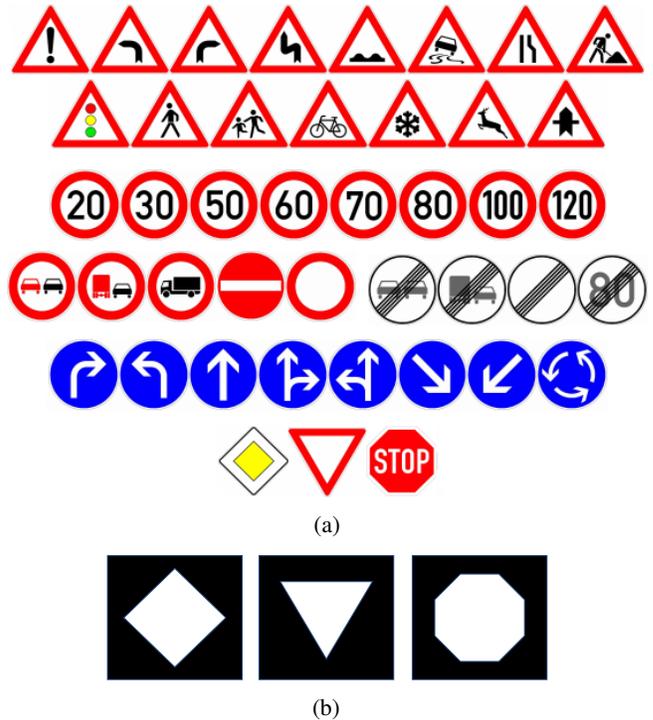

Fig. 5: An illustration of human reasoning and shape bias for mapping new samples to dataset classes. a) Representative images of GTSRB classes, and b) images of diamond, inverted triangle and octagon shapes. The task is to map images in (b) to one of the classes of GTSRB dataset. Although the color information is lost, humans can easily classify them respectively as the three images of the last row of (a). The reason is that the images from last row of (a) are the only signs with diamond, inverted triangle and octagon shapes.

### B. Security Implication

New approaches in computer vision try to understand and imitate the human visual system [31], [32]. However, it has been shown that image classification algorithms, although capable of achieving high accuracies on regular data, show certain differences with the human perception of the objects. One type of such differences is the existence of images, which are completely unrecognizable to humans, but learning models are *fooled* into classifying them as valid objects with high confidence [33]. Another type is adversarial examples [34]. That is, an adversary can deceive image classifiers by slightly modifying input images, for which a human observer would recognize the correct objects.

In this paper, we showed the fragility of CNNs against transformed inputs. It implies that, an adversary, who does not have any access to the machine learning model, can easily deceive it by applying transformations to the input images, which do not affect the human perception. We call such transformations as *adversarial transformations* and the transformed inputs as *semantic adversarial examples*.

Note that for generating regular adversarial examples, the perturbation must be small to remain unnoticeable by a human observer [34]. In contrast, adversarial transformation may introduce a large perturbation to the image, yet a human observer would correctly recognize it. We presented image complementing as one such transformation. Other kinds of adversarial transformations can be thought to be changing the color, size, brightness or orientation of the object.

## VI. CONCLUSION

In this paper we showed that, despite the impressive performance of CNNs on images distributed similar to training data, their accuracy is much lower on negative images. Our observations indicate that CNNs that are simply trained on raw data perform poorly in recognizing the semantics of objects. We also introduced the notion of semantic adversarial examples as transformed inputs which appear the same to a human observer, yet the machine learning model does not classify them correctly.

## REFERENCES

[1] D. C. Cireşan, U. Meier, L. M. Gambardella, and J. Schmidhuber, "Deep, big, simple neural nets for handwritten digit recognition," *Neural computation*, vol. 22, no. 12, pp. 3207–3220, 2010.
[2] A. Krizhevsky, I. Sutskever, and G. E. Hinton, "Imagenet classification with deep convolutional neural networks," in *Advances in neural information processing systems*, 2012, pp. 1097–1105.
[3] K. Simonyan and A. Zisserman, "Very deep convolutional networks for large-scale image recognition," *arXiv preprint arXiv:1409.1556*, 2014.
[4] L. Wan, M. Zeiler, S. Zhang, Y. L. Cun, and R. Fergus, "Regularization of neural networks using dropconnect," in *Proceedings of the 30th International Conference on Machine Learning (ICML-13)*, 2013, pp. 1058–1066.
[5] J. T. Springenberg, A. Dosovitskiy, T. Brox, and M. Riedmiller, "Striving for simplicity: The all convolutional net," *arXiv preprint arXiv:1412.6806*, 2014.
[6] C.-Y. Lee, S. Xie, P. W. Gallagher, Z. Zhang, and Z. Tu, "Deeply-supervised nets," in *AISTATS*, vol. 2, no. 3, 2015, p. 5.
[7] I. Goodfellow, J. Pouget-Abadie, M. Mirza, B. Xu, D. Warde-Farley, S. Ozair, A. Courville, and Y. Bengio, "Generative adversarial nets," in *Advances in neural information processing systems*, 2014, pp. 2672–2680.
[8] A. Dosovitskiy, J. Tobias Springenberg, and T. Brox, "Learning to generate chairs with convolutional neural networks," in *Proceedings of the IEEE Conference on Computer Vision and Pattern Recognition*, 2015, pp. 1538–1546.
[9] S. Ritter, D. G. Barrett, A. Santoro, and M. M. Botvinick, "Cognitive psychology for deep neural networks: A shape bias case study," *arXiv preprint arXiv:1706.08606*, 2017.
[10] Y. LeCun, L. Bottou, Y. Bengio, and P. Haffner, "Gradient-based learning applied to document recognition," *Proceedings of the IEEE*, vol. 86, no. 11, pp. 2278–2324, 1998.
[11] C. Zhang, S. Bengio, M. Hardt, B. Recht, and O. Vinyals, "Understanding deep learning requires rethinking generalization," *arXiv preprint arXiv:1611.03530*, 2016.
[12] M. Paulin, J. Revaud, Z. Harchaoui, F. Perronnin, and C. Schmid, "Transformation pursuit for image classification," in *Proceedings of the IEEE Conference on Computer Vision and Pattern Recognition*, 2014, pp. 3646–3653.
[13] K. Chatfield, K. Simonyan, A. Vedaldi, and A. Zisserman, "Return of the devil in the details: Delving deep into convolutional nets," *arXiv preprint arXiv:1405.3531*, 2014.
[14] S. Hauberg, O. Freifeld, A. B. L. Larsen, J. Fisher, and L. Hansen, "Dreaming more data: Class-dependent distributions over diffeomorphisms for learned data augmentation," in *Artificial Intelligence and Statistics*, 2016, pp. 342–350.
[15] J. Yosinski, J. Clune, Y. Bengio, and H. Lipson, "How transferable are features in deep neural networks?" in *Advances in neural information processing systems*, 2014, pp. 3320–3328.
[16] L. Shao, F. Zhu, and X. Li, "Transfer learning for visual categorization: A survey," *IEEE transactions on neural networks and learning systems*, vol. 26, no. 5, pp. 1019–1034, 2015.
[17] H.-C. Shin, H. R. Roth, M. Gao, L. Lu, Z. Xu, I. Nogues, J. Yao, D. Mollura, and R. M. Summers, "Deep convolutional neural networks for computer-aided detection: Cnn architectures, dataset characteristics and transfer learning," *IEEE transactions on medical imaging*, vol. 35, no. 5, pp. 1285–1298, 2016.
[18] K. Saenko, B. Kulis, M. Fritz, and T. Darrell, "Adapting visual category models to new domains," *Computer Vision–ECCV 2010*, pp. 213–226, 2010.
[19] R. Gopalan, R. Li, and R. Chellappa, "Domain adaptation for object recognition: An unsupervised approach," in *Computer Vision (ICCV), 2011 IEEE International Conference on*. IEEE, 2011, pp. 999–1006.
[20] J. Donahue, Y. Jia, O. Vinyals, J. Hoffman, N. Zhang, E. Tzeng, and T. Darrell, "Decaf: A deep convolutional activation feature for generic visual recognition," in *International conference on machine learning*, 2014, pp. 647–655.
[21] S. J. Pan and Q. Yang, "A survey on transfer learning," *IEEE Transactions on knowledge and data engineering*, vol. 22, no. 10, pp. 1345–1359, 2010.
[22] Y. LeCun, C. Cortes, and C. J. Burges, "The mnist database of handwritten digits," 1998.
[23] A. Krizhevsky, "Learning multiple layers of features from tiny images," 2009.
[24] J. Stallkamp, M. Schlipsing, J. Salmen, and C. Igel, "Man vs. computer: Benchmarking machine learning algorithms for traffic sign recognition," *Neural networks*, vol. 32, pp. 323–332, 2012.
[25] O. Russakovsky, J. Deng, H. Su, J. Krause, S. Satheesh, S. Ma, Z. Huang, A. Karpathy, A. Khosla, M. Bernstein *et al.*, "Imagenet large scale visual recognition challenge," *International Journal of Computer Vision*, vol. 115, no. 3, pp. 211–252, 2015.
[26] O. Vinyals, C. Blundell, T. Lillicrap, D. Wierstra *et al.*, "Matching networks for one shot learning," in *Advances in Neural Information Processing Systems*, 2016, pp. 3630–3638.
[27] P. Tabacof and E. Valle, "Exploring the space of adversarial images," in *Neural Networks (IJCNN), 2016 International Joint Conference on*. IEEE, 2016, pp. 426–433.
[28] L. Fei-Fei, R. Fergus, and P. Perona, "One-shot learning of object categories," *IEEE transactions on pattern analysis and machine intelligence*, vol. 28, no. 4, pp. 594–611, 2006.
[29] R. Socher, M. Ganjoo, C. D. Manning, and A. Ng, "Zero-shot learning through cross-modal transfer," in *Advances in neural information processing systems*, 2013, pp. 935–943.
[30] A. Santoro, S. Bartunov, M. Botvinick, D. Wierstra, and T. Lillicrap, "One-shot learning with memory-augmented neural networks," *arXiv preprint arXiv:1605.06065*, 2016.
[31] J. J. DiCarlo, D. Zoccolan, and N. C. Rust, "How does the brain solve visual object recognition?" *Neuron*, vol. 73, no. 3, pp. 415–434, 2012.
[32] N. Kruger, P. Janssen, S. Kalkan, M. Lappe, A. Leonardis, J. Piater, A. J. Rodriguez-Sanchez, and L. Wiskott, "Deep hierarchies in the primate visual cortex: What can we learn for computer vision?" *IEEE transactions on pattern analysis and machine intelligence*, vol. 35, no. 8, pp. 1847–1871, 2013.
[33] A. Nguyen, J. Yosinski, and J. Clune, "Deep neural networks are easily fooled: High confidence predictions for unrecognizable images," in *Proceedings of the IEEE Conference on Computer Vision and Pattern Recognition*, 2015, pp. 427–436.
[34] C. Szegedy, W. Zaremba, I. Sutskever, J. Bruna, D. Erhan, I. Goodfellow, and R. Fergus, "Intriguing properties of neural networks," *arXiv preprint arXiv:1312.6199*, 2013.